\documentclass{SCIS2026}

\begin{document}
\newcommand{\best}[1]{\textbf{#1}}
\newcommand{\second}[1]{\underline{#1}}
\newcommand{\tabfs}{\footnotesize}
\newcommand{\tabstretch}{\renewcommand{\arraystretch}{1.05}}
\ArticleType{RESEARCH PAPER}
\Year{2025}
\Month{January}
\Vol{68}
\No{1}
\DOI{}
\ArtNo{}
\ReceiveDate{}
\ReviseDate{}
\AcceptDate{}
\OnlineDate{}
\AuthorMark{}
\AuthorCitation{}

\title{PhysMLLMs: Spatial Priors for Unified Referring Segmentation and Grounded Reasoning of Images and Videos}{Title for citation}

\author[1]{Siyao Yan}{}
\author[1]{Bo Han}{}
\author[1]{Jisheng Dang}{{dangjisheng@lzu.edu.cn}}
\author[1]{Bimei Wang}{} 
\author[1]{Shude Wang}{}
\author[1]{Hong Peng}{{pengh@lzu.edu.cn}}
\author[2]{\\Yulan Guo}{}
\author[2]{Jianhuang Lai}{}
\author[1]{Bin Hu}{{bh@lzu.edu.cn}}
\author[3]{Tat-SengChua}{}


\address[1]{School of Information Science and Engineering, Lanzhou University, Lanzhou, 730000, China}
\address[2]{School of Electronics and Communication Engineering, Sun Yat-sen University, Shenzhen 510006, China}
\address[3]{School of Computing, National University of Singapore, Singapore, 119077, Singapore }

\abstract{Video multimodal large language models support language guided video segmentation, but they often show spatio temporal inconsistencies, e.g., jitter, drift, and identity switches. These failures are more common when targets are partly hidden or when similar objects appear nearby.One likely reason is that current training lacks explicit spatial priors, which makes it difficult to maintain stable spatial identity and shape over time.
We present PhysMLLMs, a training-stage prior injection architecture that injects physics-inspired spatial continuity priors into Video MLLMs. PhysMLLMs is designed to encourage more stable object-centered representations by aligning the student global visual representation with a frozen teacher model during training. Our core mechanism, Global Representation Prior Alignment (REPA-Global), distills global visual representations from a frozen DINOv2 teacher using an offline embedding cache and a scheduled distillation plan. This design keeps inference unchanged and does not add inference time cost.
Across multiple video benchmarks, PhysMLLMs improves video segmentation mask quality and cross-frame consistency, with larger gains on challenging cases involving small targets, fast motion, occlusion, distractors, and reasoning queries. On single-frame referring image segmentation and representative general VLM benchmarks, PhysMLLMs maintains comparable performance, demonstrating that the injected spatial prior improves video consistency without compromising image-level grounding or general multimodal capability. These results suggest that physics-inspired spatial prior injection can improve temporal stability while preserving general capability. The code is available at https://github.com/tusu-code/20260121-icml2026-2.git.
}

\keywords{Language-guided video segmentation, spatiotemporal consistency, representation distillation, knowledge distillation, prior injection, parameter-efficient fine-tuning}

\maketitle

\section{Introduction}
Video multimodal large language models, or Video MLLMs, have recently shown strong semantic grounding through large scale image text alignment and instruction tuning. When combined with strong mask priors and lightweight decoders, they provide a practical backbone for language guided video segmentation \cite{liu2023llava,kirillov2023segmentanything}. However, the main training signals give limited direct guidance for keeping predictions consistent over time. As a result, masks that look reasonable in each frame can still become unstable across frames, especially under occlusion, distractors, and appearance changes.
\begin{figure*}[t]
  \centering
  \includegraphics[width=\textwidth]{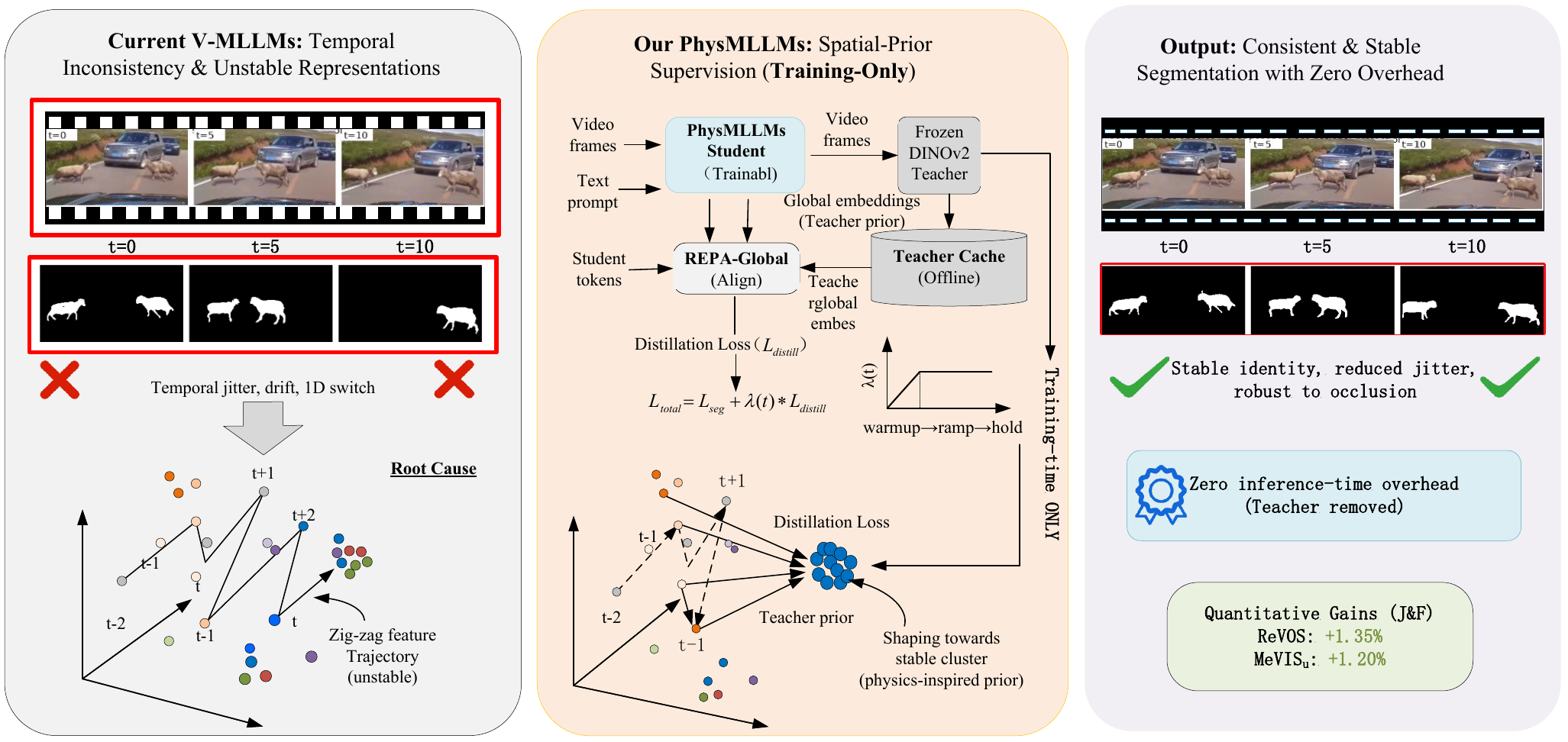}
  \caption{Motivation and overview of spatial prior injection for video segmentation. Unstable representations can lead to temporally inconsistent masks (left). We inject a continuity prior by distilling global visual representations from a frozen DINOv2 teacher using an offline teacher cache and scheduled distillation weight (middle). The resulting model yields more consistent segmentations with zero inference-time overhead, since the teacher is removed at test time (right).}
  \label{fig:motivation}
\end{figure*}
Figure \ref{fig:motivation} shows a key pattern. Many failures in video segmentation come from unstable feature changes over time rather than from isolated errors in the mask decoder. This observation motivates a training strategy that guides representation dynamics toward stable object centered features, instead of relying only on stronger heads or test time rules.

We focus on three common failure patterns in practice. They include temporal jitter, drifting in identity or shape, and identity switches after occlusion. These issues reflect missing spatial priors, which weakens spatial object permanence and spatial continuity across frames. To make this failure mode more concrete, we further examine the official Sa2VA-InternVL3-2B baseline on the ReVOS validation split. Its overall J\&F is 54.61, but the score drops to 33.65 on small-target cases and 31.29 on occlusion or disappearance cases. These decreases of 20.96 and 23.32 points show that the main difficulty is not only frame-level mask prediction, but also the preservation of stable object identity and spatial continuity under challenging video conditions. This observation motivates a training-time prior that regularizes visual representations toward temporally coherent object-centric features. We do not rely on explicit physics-based simulation or hand-crafted physical equations. Instead, we use a physics-inspired spatial consistency prior: once an object is identified, its representation should remain coherent over time unless the video provides clear evidence of a real physical change. This prior reflects object permanence, spatial continuity, and temporal coherence in the physical world. This perspective suggests that higher per frame accuracy alone is often not enough.
Memory based association is a common remedy in video segmentation \cite{oh2019stm,cheng2021stcn,cheng2022xmem}, but it may not fully fix instability when upstream Video MLLM features are not well regularized. In our experiments, unstable masks can remain even with memory, which suggests that the main bottleneck often lies in the representation space.

We propose a training-stage prior injection strategy that transfers object centered and geometry aware priors from a strong self supervised teacher, DINOv2 \cite{oquab2023dinov2}, into a Video MLLM through representation distillation \cite{hinton2015distilling}. Our core mechanism, REPA-Global, aligns the student global visual representation with cached DINOv2 global embeddings. This provides a representation-level regularizer that is expected to reduce abrupt cross-frame shifts while keeping inference unchanged. To make distillation scalable, we precompute teacher embeddings for each frame offline and distill from the cache during training, which avoids repeated teacher forward passes. To reduce interference with general multimodal capability and to stabilize training, we combine a scheduled distillation weight with calibrated parameter efficient fine tuning, including Vision LoRA \cite{hu2022lora} and MaskDecoder only calibration. This design routes most injected gradients into the visual and decoding pathway while keeping the core vision language alignment largely intact.

Our contributions are summarized below.
\begin{itemize}
\item We propose PhysMLLMs, a training-stage prior injection strategy that injects physics-inspired spatial continuity priors into Video MLLMs to improve temporal stability for language-guided video segmentation and support more stable video reasoning behavior, while preserving the general grounding ability of the underlying vision-language model.
\item We develop REPA-Global, which aligns the student global embedding with a frozen DINOv2 teacher to inject a physics-inspired spatial continuity prior that improves cross-frame consistency with no added inference-time cost.
\item We make the training strategy practical at scale by combining an offline teacher cache, a scheduled distillation plan, and a calibrated parameter efficient fine tuning recipe, which improves efficiency and reduces the risk of general capability drop during specialization.
\item Extensive experiments show that PhysMLLMs consistently improves language-guided video segmentation on ReVOS, MeVIS, and Ref-DAVIS17, especially in challenging video scenarios with small targets, fast motion, occlusion, distractors, and reasoning queries. Meanwhile, it preserves competitive single-frame referring segmentation performance on RefCOCO, RefCOCO+, and RefCOCOg, and shows no clear degradation on representative general VLM benchmarks.
\end{itemize}

\section{Related Work}

\subsection{Referring Video Object Segmentation}
Referring video object segmentation extends video object segmentation by adding language guidance. It requires accurate grounding of free form expressions and masks that remain consistent across frames. Compared with class agnostic tracking, the model must resolve language ambiguity, handle multiple similar instances, and keep the target identity stable over time. Common evaluation suites include DAVIS \cite{ponttuset2017davis} and YouTube VOS \cite{xu2018youtubevos}, as well as language guided datasets such as Ref YouTube VOS \cite{seo2022refyoutubevos} and MeVIS \cite{ding2023mevis}. These benchmarks cover diverse scenes and motion patterns, and they often stress occlusion, viewpoint change, and appearance variation, which can expose long range instability.

Earlier systems often improve temporal association with memory based designs, e.g., STM \cite{oh2019stm}, STCN \cite{cheng2021stcn}, and XMem \cite{cheng2022xmem}. Memory helps reuse past cues and reduce short term flicker, but it can still struggle when the upstream representation drifts or when the query target is visually similar to distractors. More recently, strong mask priors and large vision language models have become widely used. Segment Anything provides open world mask priors \cite{kirillov2023segmentanything}, and vision language models, e.g., LLaVA, improve semantic grounding \cite{liu2023llava}. In practice, these components can improve per frame mask quality and language alignment, yet long range stability remains difficult under occlusion and appearance changes. Predictions may show jitter, drift, and identity switches over time. Our PhysMLLMs architecture addresses these consistency problems at the representation level and uses a training stage prior injection strategy to improve temporal stability in video segmentation. Classical memory-based video object segmentation methods, such as STCN and XMem, provide strong temporal association when an object mask or reliable object cue is available. However, they are not directly comparable to language-guided Video-MLLM systems because they do not resolve free-form referring expressions or generate masks from open-ended language queries. We therefore discuss them as representative temporal association methods, while the main quantitative comparison focuses on language-guided Video-MLLM systems that share a closer input-output protocol.

\subsection{Self-Supervised Spatial Priors and Geometric Structure }
Self supervised pretraining can learn transferable visual priors from large scale unlabeled data. These priors are useful as teacher signals when dense video mask labels are limited and when the training distribution is diverse. Contrastive and bootstrap methods learn invariances and object centered cues \cite{chen2020simclr,he2020moco,grill2020byol,caron2020swav}. Masked prediction methods provide complementary supervision for dense representations \cite{he2022maskedautoencoders,bao2021beit,zhou2021ibot,wei2021maskfeat}. DINO style self distillation learns strong semantic features and often improves sensitivity to shape and geometry \cite{caron2021dino}. DINOv2 further strengthens these properties through scaling and data curation \cite{oquab2023dinov2}. These properties make such teachers attractive for transferring stable global structure without requiring extra manual labels.

For video, masked autoencoding, e.g., VideoMAE, naturally uses temporal structure and can capture motion related cues \cite{tong2022videomae}. Other self supervised video objectives also encourage temporal invariance, which can support smoother feature evolution even when the pixel content changes quickly. These priors provide a practical way to encourage temporally coherent features beyond frame by frame segmentation supervision. They also relate to spatial awareness because stable object identity over time depends on spatial continuity priors that standard segmentation losses do not always enforce, especially when the target becomes small, blurred, or partly hidden.

\subsection{Consistency Oriented Distillation and Representation Alignment}
Knowledge distillation has moved beyond logit matching and now often uses feature level alignment and geometry preserving objectives. In this setting, the teacher provides structural guidance for student representations \cite{Gou_2021}. Supervision on intermediate features can provide steadier signals and improve spatial selectivity \cite{romero2015fitnetshintsdeepnets}. Relation aware and contrastive variants can also preserve the geometry of the feature space \cite{Park_2019_CVPR}. These methods are often useful when the task needs stable structure and when the student model must retain general ability while specializing.

For dense prediction, structured distillation can be more reliable than direct pixel matching when global layout and long range dependencies matter \cite{liu2020structuredknowledgedistillationdense}. Global alignment can also reduce sensitivity to local noise, which is important when distractors trigger large false masks. These findings motivate our REPA-Global design within the PhysMLLMs architecture. We align global visual representations to transfer object centered priors that support persistent identity and coherent geometry over time \cite{pmlr-v139-touvron21a}. Because overly strong alignment may suppress valid changes, we treat the distillation weight as a training control variable rather than a fixed constant \cite{han2024parameterefficientfinetuninglargemodels}. This design supports physics-inspired spatial continuity priors, such as object permanence and temporal coherence, while avoiding an overly rigid representation.

\subsection{Parameter Efficient Fine Tuning and Capability Preservation}
Parameter efficient fine tuning adapts pretrained models while limiting representational drift, which helps preserve multimodal abilities \cite{pmlr-v97-houlsby19a}. This is important for Video MLLMs because the same model is often expected to support both segmentation and general language based reasoning. Adapter based methods and composition strategies can reduce forgetting and support safer task updates \cite{pfeiffer2021adapterfusionnondestructivetaskcomposition}. Low rank and prompt based tuning further constrains the learnable space and often improves efficiency with reduced interference \cite{hu2022lora,li-liang-2021-prefix,lester-etal-2021-power}. These approaches also make it easier to control where the model changes, which matters when adding new training signals.

In our setting, Vision LoRA and MaskDecoder only calibration act as practical guardrails. They direct most injected gradients into the visual and decoding pathways and reduce unnecessary changes to vision language alignment. Capacity remains an important factor. If the parameter efficient budget is too small, the model may not absorb the injected priors well, especially when the training data contains long occlusions or fast motion. This makes module placement and rank important for performance.

\section{Method}

\begin{figure*}[!t]
  \centering
  \includegraphics[width=\textwidth]{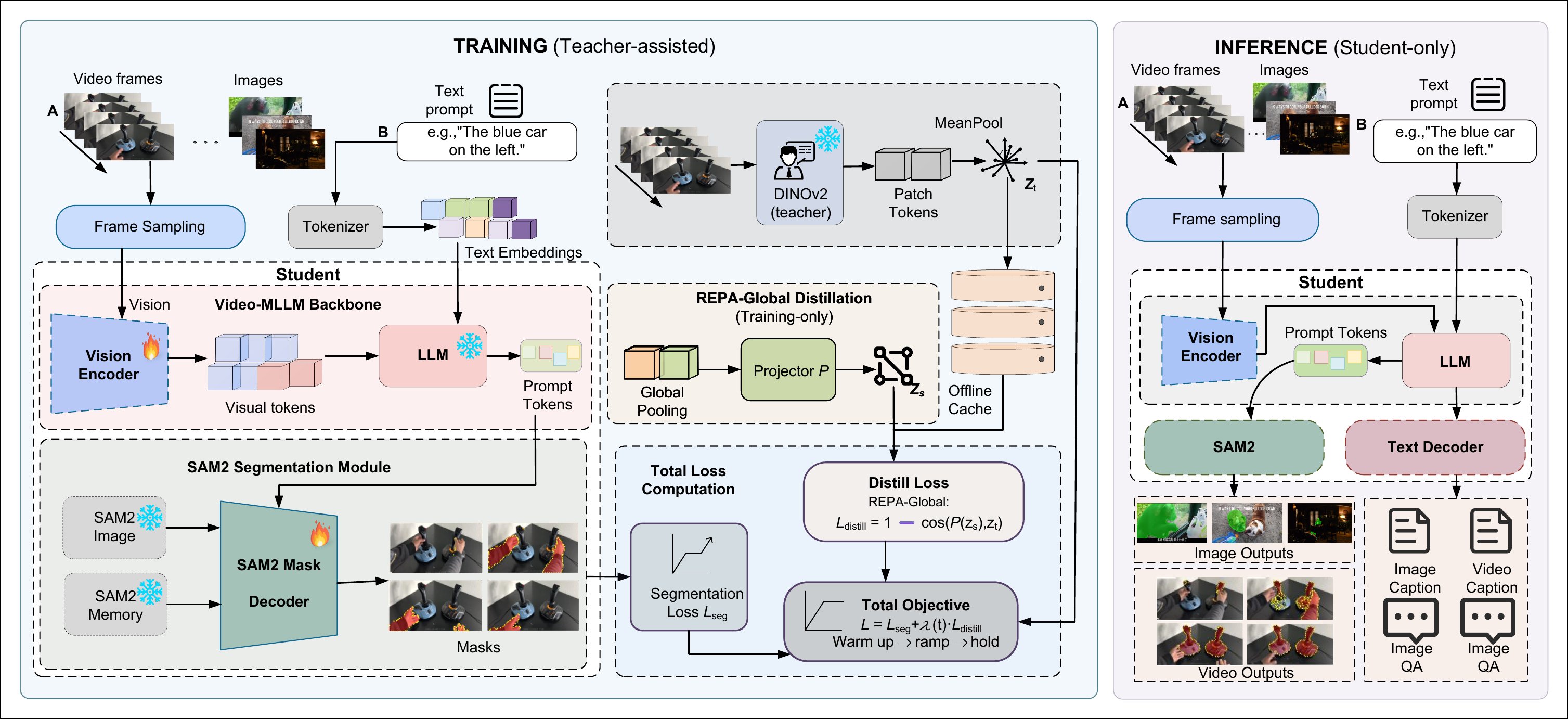}
  \caption{PhysMLLMs overview. During training, REPA-Global injects a physics-inspired spatial continuity prior by aligning the student global visual embedding $\mathbf{z}_{\mathrm{s}}$ with a frozen DINOv2 teacher embedding $\mathbf{z}_{\mathrm{t}}$ retrieved from an offline cache. Text embeddings condition the LLM and downstream mask prediction, but they do not enter the global visual pooling branch before teacher alignment. At inference, the teacher branch and cache are removed.}
  \label{fig:method-overview}
\end{figure*}
\subsection{Overview}
We present PhysMLLMs, a transferable prior injection architecture for large video segmentation models. PhysMLLMs uses a training stage strategy that injects spatially grounded consistency through representation distillation. Our goal is not to rely on test time heuristics, e.g., optical flow or post processing. Instead, we guide how visual representations evolve over time during training. This design supports deployment because the teacher is used only during training and is removed at inference, so runtime cost remains unchanged, as shown in Fig.\ref{fig:method-overview}.

In this work, the term physics-inspired prior refers to a soft constraint derived from object permanence, spatial continuity, and temporal coherence, rather than an explicit physical simulator. We implement this prior through representation-level distillation, so that the student representation is encouraged to evolve smoothly across frames while still allowing real appearance and pose changes.
Given a training sample, the student predicts segmentation masks and we optimize it with the standard segmentation loss $\mathcal{L}_{\mathrm{seg}}$. We further add a representation-level distillation term $\mathcal{L}_{\mathrm{distill}}$ to encode spatial priors, e.g., spatial persistence and geometric continuity of the target. We observe that temporally implausible outputs, e.g., jitter, drift, and identity switches after occlusion, can happen even when each frame looks locally correct. This often correlates with unstable feature changes across frames. We therefore impose a consistency constraint in representation space rather than using pixel level smoothing, which can depend strongly on the decoder design.

The overall objective is
\begin{equation}
\mathcal{L} = \mathcal{L}_{\mathrm{seg}} + \lambda(t)\mathcal{L}_{\mathrm{distill}},
\label{eq:overall_loss}
\end{equation}
where $t$ denotes the training step and $\lambda(t)$ controls the strength of the injected prior. We treat $\lambda(t)$ as a training control signal rather than a fixed scalar. If the constraint becomes too strong too early, it can weaken grounding and localization and may lead to over smoothing. We therefore use a progressive schedule as part of the training strategy, described in Sec. \ref{sec:stability}.
We implement the student on Sa2VA\cite{yuan2025sa2va}. A pretrained multimodal backbone produces visual tokens, and a segmentation decoder predicts masks conditioned on the language query. For each frame $x$, we extract student visual tokens $\mathbf{S}(x)$ from the visual encoder, pool them into a student global embedding in Eq. \ref{eq:student_global_raw}, and map it through a lightweight projector for teacher alignment in Sec. \ref{sec:repa-global}. We compute the distillation loss per frame and aggregate it over frames within a batch. This provides a representation level regularizer that complements the mask loss.

\subsection{REPA-Global Distillation}
\label{sec:repa-global}
We instantiate $\mathcal{L}_{\mathrm{distill}}$ with a global prior alignment term, denoted as REPA-Global. We use DINOv2 \cite{oquab2023dinov2} with ViT B 14 as the frozen vision teacher. For an input frame $x$, the teacher produces patch tokens
\begin{equation}
\mathbf{T}(x)\in\mathbb{R}^{N\times D}, \quad D=768,
\label{eq:teacher_tokens}
\end{equation}
and we compute the teacher global embedding by mean pooling
\begin{equation}
\mathbf{z}_{\mathrm{t}}(x) = \mathrm{MeanPool}\!\left(\mathbf{T}(x)\right)\in\mathbb{R}^{D}.
\label{eq:teacher_global}
\end{equation}

The student provides visual tokens
\begin{equation}
\mathbf{S}(x)\in\mathbb{R}^{M\times C},
\label{eq:student_tokens}
\end{equation}
and we obtain a student global embedding by pooling
\begin{equation}
\mathbf{z}_{\mathrm{s}}(x)=\mathrm{Pool}\!\left(\mathbf{S}(x)\right)\in\mathbb{R}^{C}.
\label{eq:student_global_raw}
\end{equation}
The student embedding $\mathbf{z}_{\mathrm{s}}(x)$ is extracted from visual tokens before text-conditioned mask prediction. Text conditioning occurs later through the LLM-generated segmentation token and the language-conditioned mask decoder. Thus, REPA-Global injects a training-time visual prior rather than distilling a text-conditioned multimodal hidden state. We use global alignment because the targeted failures, such as jitter, drift, and identity switches, often reflect unstable object-level trajectories rather than isolated patch errors. Dense token-level imitation may also over-constrain query-irrelevant regions, so we treat token-level alignment as an ablation.
Our default choice is mean pooling over the visual tokens, and we treat CLS pooling as an optional variant. Since $C\neq D$, we use a lightweight projector $P(\cdot)$, implemented as a linear layer, to map the student embedding to the teacher dimension. We then align normalized embeddings using cosine distance:
\begin{equation}
\mathcal{L}^{\mathrm{global}}_{\mathrm{distill}}
=
\mathbb{E}_{x}
\left[
1-
\cos
\left(
\widehat{P(\mathbf{z}_{\mathrm{s}}(x))},
\widehat{\mathbf{z}_{\mathrm{t}}(x)}
\right)
\right],
\label{eq:global_distill}
\end{equation}
where both $P(\mathbf{z}_{\mathrm{s}}(x))$ and $\mathbf{z}_{\mathrm{t}}(x)$ are L2-normalized before computing the cosine distance. We use cosine alignment for scale robustness, which often improves optimization stability.

As an optional variant, we also evaluate MSE on normalized embeddings. REPA-Global provides a training-time representation-level regularization signal that encourages physics-inspired object-centered features without tying the training strategy to a specific mask head.

The novelty is not the use of DINOv2 alone, but its integration with a cache-enabled training workflow, scheduled prior injection, and calibrated PEFT. This makes the teacher a training-only physics-inspired prior provider while keeping the inference graph identical to the original student.

\subsection{Calibrated PEFT for Prior Injection}
Injecting priors through distillation can interact with multimodal alignment. To limit unwanted drift and keep deployment simple, we use a parameter efficient tuning strategy. We freeze the language model and most pretrained modules, and we update only two parts. We enable Vision LoRA \cite{hu2022lora} in the visual backbone and we calibrate only the mask decoder on the segmentation side.

For a selected linear weight matrix $\mathbf{W}_0$, LoRA parameterizes the update as
\begin{equation}
\mathbf{W}=\mathbf{W}_0+\Delta\mathbf{W},
\quad
\Delta\mathbf{W}=\mathbf{B}\mathbf{A}.
\label{eq:lora}
\end{equation}
In this setup, the base matrix $\mathbf{W}_0$ stays frozen and the model learns only the low rank factors $\mathbf{A}$ and $\mathbf{B}$. This constrained update helps preserve semantic grounding while allowing the injected prior to shape visual features that affect temporal stability. A practical limitation is capacity. If the parameter efficient budget is too small, the model may not absorb the spatial priors well, so module placement and rank become important hyperparameters.

\subsection{Overall Training Workflow for Cache, Scheduling, and Deployment}
\label{sec:stability}

Stable prior injection requires consistent teacher supervision and controlled regularization strength. As summarized in Algorithm~\ref{alg:end2end}, we first precompute the teacher global embedding $\mathbf{z}_{\mathrm{t}}(x)$ for each training frame and store it in an offline cache. During training, the student reads cached teacher embeddings to compute $\mathcal{L}_{\mathrm{distill}}$, which avoids repeated teacher forward passes and keeps the distillation target deterministic.

We treat $\lambda(t)$ as a control variable and use a warmup-ramp-hold schedule:
\begin{equation}
\lambda(t)=
\begin{cases}
0, & t < t_{\mathrm{w}},\\
\lambda_{\max}\cdot \frac{t-t_{\mathrm{w}}}{t_{\mathrm{r}}-t_{\mathrm{w}}}, & t_{\mathrm{w}}\le t < t_{\mathrm{r}},\\
\lambda_{\max}, & t\ge t_{\mathrm{r}},
\end{cases}
\label{eq:schedule}
\end{equation}
where $t_{\mathrm{w}}$ and $t_{\mathrm{r}}$ denote the end of warmup and ramp. The warmup stage optimizes only the segmentation objective, while the scheduled stage introduces the physics-inspired prior through $\mathcal{L}_{\mathrm{distill}}$ and updates only calibrated PEFT parameters. At inference, the teacher and cache are removed, and only the student model is used.

\begin{algorithm}[!t]
\caption{End-to-end workflow for cache-enabled prior injection}
\label{alg:end2end}
\begin{algorithmic}[1]
\REQUIRE Training frames $\{x\}$; frozen teacher $\mathbf{T}$; student $\mathbf{S}_{\theta}$; schedule $\lambda(t)$; max weight $\lambda_{\max}$
\ENSURE Trained student parameters $\theta$ for deployment

\STATE \textbf{Cache generation (offline)}
\FORALL{training frame $x$}
  \STATE compute teacher global embedding $\mathbf{z}_{\mathrm{t}}(x)$
  \STATE store $\mathbf{z}_{\mathrm{t}}(x)$ in the teacher cache
\ENDFOR

\STATE \textbf{Warmup}
\FOR{training step $t < t_{\mathrm{w}}$}
  \STATE set $\lambda(t) \leftarrow 0$
  \STATE update $\theta$ by minimizing $\mathcal{L}_{\mathrm{seg}}$
\ENDFOR

\STATE \textbf{Ramp and hold}
\FOR{training step $t \ge t_{\mathrm{w}}$}
  \STATE read cached $\mathbf{z}_{\mathrm{t}}(x)$ for frames in the current batch
  \STATE compute $\mathcal{L}_{\mathrm{distill}}$
  \STATE optimize $\mathcal{L}_{\mathrm{seg}}+\lambda(t)\mathcal{L}_{\mathrm{distill}}$
  \STATE update only calibrated PEFT parameters
\ENDFOR

\STATE \textbf{Deployment}
\STATE discard the teacher and cache; run inference with the student only
\end{algorithmic}
\end{algorithm}

\section{Experiments}

We evaluate whether PhysMLLMs improves language-guided video segmentation while preserving image-level grounding and general VLM capability. We first clarify the setup and evaluation protocol, then compare with recent Video-MLLMs, analyze the effect of prior injection, and summarize capability-preservation results.

\subsection{Experimental Setup and Evaluation Protocol}
\label{sec:experimental_setup}

Unless otherwise stated, PhysMLLMs is built on Sa2VA-InternVL3-2B, which combines an InternVL3-2B multimodal backbone with the SAM2 segmentation module. The full student model contains approximately 2B parameters. All controlled comparisons and ablations use this same 2B backbone, so the observed differences come from REPA-Global rather than model scale. Table~\ref{tab:recent_videomllm} includes Sa2VA-8B only as an external reference.

We use a frozen DINOv2 ViT-B/14 teacher and precompute frame-level teacher embeddings into an offline cache. During inference, the teacher and cache are removed. To preserve vision-language alignment, we freeze the language model and most pretrained modules, and update only Vision-LoRA and mask-decoder calibration parameters. The settings \texttt{dw008}, \texttt{dw012}, and \texttt{dw020} denote peak distillation weights $\lambda_{\max}=0.008$, $0.012$, and $0.020$ under the same warmup-ramp-hold schedule.

For video segmentation, we report DAVIS-style $\mathcal{J}$, $\mathcal{F}$, and $\mathcal{J\&F}$ when available, and use $\mathcal{J\&F}$ as the main metric for cross-method comparison because many published Video-MLLM systems do not report separate $\mathcal{J}$ and $\mathcal{F}$. For image grounding, we report RefCOCO-series accuracy as a non-degradation check. For general VLM capability, we evaluate MMBench, MME, POPE, and TextVQA, with detailed results provided in the appendix.

\subsection{Main Comparison with Recent Video-MLLM Systems}
\label{sec:main_comparison}

Table~\ref{tab:recent_videomllm} positions PhysMLLMs relative to representative recent Video-MLLM systems on MeVIS U, ReVOS, and Ref-DAVIS17, when the corresponding numbers are available. To reflect recent progress, we include recently reported Video-MLLM segmentation systems such as HyperSeg-3B and InstructSeg when their published results are available. The comparison covers a range of model sizes and training strategies, and it includes the Sa2VA baseline that our method builds upon. PhysMLLMs improves consistently across the reported benchmarks, indicating that representation-level prior injection is not tied to a single dataset or evaluation setup.

It is worth noting that Table~\ref{tab:recent_videomllm} contains models with different backbone capacities, including Sa2VA-8B and Sa2VA-InternVL3-2B. Therefore, this table is used to position PhysMLLMs among recent Video-MLLM systems, rather than to serve as the only evidence for a controlled capacity-matched comparison. The controlled comparisons are conducted under the same Sa2VA-InternVL3-2B + SAM2 backbone in the following ablation and capability-check analyses.Some compared methods report results on only one or two datasets. We keep these entries as ``--'' rather than re-estimating them under unmatched settings, because differences in model size, training data, and evaluation protocol may otherwise confound the comparison. This reporting follows the published availability of each method and avoids introducing non-comparable numbers.

We use the aggregate J\&F score in Table~\ref{tab:recent_videomllm} to maximize coverage across published Video-MLLM systems. Separate $\mathcal{J}$ and $\mathcal{F}$ values are reported in controlled internal analyses when they are available, such as Table~\ref{tab:global_token_ablation}.

\begin{table}[!t]
\centering
\tabfs
\tabstretch
\setlength{\tabcolsep}{4.6pt}
\caption{
Comparison with recent Video-MLLM systems on language-guided video segmentation. We report DAVIS-style J\&F on MeVIS U, ReVOS, and Ref-DAVIS17 because most published Video-MLLM papers report aggregate J\&F but do not consistently provide separate $\mathcal{J}$ and $\mathcal{F}$ values across all datasets. Higher is better. ``--'' means not reported or not available under a comparable protocol.\label{tab:recent_videomllm}}

\begin{tabular}{lccc}
\toprule
Method & MeVIS U & ReVOS & Ref-DAVIS17 \\
\midrule
PG-Video-LLaVA~\cite{munasinghe2023pgvideollava} & 18.9 & -- & -- \\
GLaMM + SAM2~\cite{munasinghe2024videoglamm} & 38.7 & -- & -- \\
VideoGLaMM-3.8B~\cite{munasinghe2024videoglamm} & 45.2 & -- & -- \\
VISA-13B~\cite{yan2024visa} & 44.5 & 50.9 & 70.4 \\
VideoLISA-3B~\cite{bai2024onetokentoseg} & 44.4 & -- & 68.8 \\
HyperSeg-3B~\cite{wei2025hyperseg} & -- & 55.7 & 71.2 \\
InstructSeg~\cite{wei2024instructseg} & -- & 54.5 & 71.1 \\
ViLLa-InternVideo2-6B~\cite{zheng2025villa} & 49.4 & \second{57.0} & 74.3 \\
GLUS-7B/GLUS-ED~\cite{lin2025glus} & 51.3 & 54.9 & -- \\
Sa2VA-8B~\cite{yuan2025sa2va} & 46.2 & 53.6 & 73.8 \\
Sa2VA-InternVL3-2B~\cite{yuan2025sa2va} & \second{53.9} & 56.2 & \second{74.5} \\
\midrule
PhysMLLMs \texttt{dw012} & \best{55.1} & \best{57.4} & \best{76.0} \\
\bottomrule
\end{tabular}
\end{table}

\begin{table}[!t]
\centering
\tabfs
\tabstretch
\setlength{\tabcolsep}{4.6pt}
\caption{
Ablation of global and token-level DINOv2 alignment on ReVOS under the same Sa2VA-InternVL3-2B + SAM2 backbone. Global-only REPA-Global outperforms token-only and global+token variants, suggesting that dense patch-level constraints may interfere with language-conditioned target selection.
}
\label{tab:global_token_ablation}
\begin{tabular}{lccc}
\toprule
Setting & ReVOS $\mathcal{J}$ & ReVOS $\mathcal{F}$ & ReVOS $\mathcal{J\&F}$ \\
\midrule
Global-only \texttt{dw012} & \best{54.31} & \best{60.51} & \best{57.41} \\
DINOv2 token-only & \second{53.32} & \second{59.21} & \second{56.26} \\
DINOv2 global+token & 53.01 & 59.12 & 56.07 \\
\bottomrule
\end{tabular}
\end{table}

\subsection{Effect of Prior Injection}
\label{sec:prior_analysis}

\textbf{Teacher-prior injection is the dominant source of gain.}
Table~\ref{tab:kd_ablation} isolates the effect of the teacher spatial prior by toggling distillation while keeping the rest of the training recipe unchanged. Across the tested settings, enabling distillation improves segmentation on ReVOS and transfers to MeVIS U. This pattern suggests that the gains are driven by the injected physics-inspired spatial prior rather than incidental optimization effects. The improvements on MeVIS U are particularly informative because this split differs from the training distribution, indicating that the continuity constraint strengthens generalization in addition to improving in-domain performance.
\begin{table}[!t]
\centering
\tabfs
\tabstretch
\setlength{\tabcolsep}{5pt}
\caption{
Ablation of teacher-prior injection under the same Sa2VA-InternVL3-2B + SAM2 backbone. The symbols \texttt{dw008}, \texttt{dw012}, and \texttt{dw020} denote peak distillation weights $\lambda_{\max}=0.008$, $0.012$, and $0.020$ under the same warmup-ramp-hold schedule. Higher J\&F is better.
}
\label{tab:kd_ablation}
\begin{tabular}{lccc}
\toprule
Setting & KD & ReVOS J\&F & MeVIS U J\&F \\
\midrule
baseline & w/o & 56.06 & 53.92 \\
\midrule
\texttt{dw008} & w/  & \second{56.88} & 54.51 \\
\texttt{dw008} & w/o & 56.46 & 54.42 \\
\texttt{dw012} & w/  & \best{57.41} & \second{55.12} \\
\texttt{dw012} & w/o & 56.76 & 54.45 \\
\texttt{dw020} & w/  & 56.87 & \best{55.24} \\
\texttt{dw020} & w/o & 56.66 & 54.32 \\
\bottomrule
\end{tabular}
\end{table}

\textbf{The gains are concentrated in difficult video scenarios.}
To better understand why the gains are more pronounced in video segmentation than in single-frame referring tasks, we conduct a complexity-stratified analysis on ReVOS. As shown in Table~\ref{tab:revos_stratified}, PhysMLLMs improves over the official Sa2VA-InternVL3-2B model by +2.26 J\&F on small targets, +1.92 on occlusion or disappearance cases, +2.02 on distractor-heavy cases, +1.37 on fast-motion cases, +1.76 on reasoning queries, and +1.59 on the hard union subset. These scenarios directly stress temporal identity preservation, cross-frame spatial continuity, and robustness to visual distractors. In contrast, RefCOCO is a single-frame benchmark where the official Sa2VA-InternVL3-2B baseline is already strong and no temporal identity maintenance is required. We therefore use the RefCOCO-series results mainly as non-degradation checks for image-level grounding, rather than as the primary target of the proposed video-oriented prior. The remaining gap on small-target and occlusion-heavy cases also suggests that extremely small or persistently occluded objects remain challenging, which clarifies the current applicability boundary of PhysMLLMs.
\begin{table}[!t]
\centering
\tabfs
\tabstretch
\setlength{\tabcolsep}{4.2pt}
\caption{
Complexity-stratified analysis on ReVOS. Here, $n$ denotes the number of ReVOS validation expressions in each stratum, rather than videos or frames. We report the official Sa2VA-InternVL3-2B result, the PhysMLLMs \texttt{dw012} result, and the absolute J\&F improvement. ``Hard union'' includes expressions satisfying at least one challenging condition.
}
\label{tab:revos_stratified}
\begin{tabular}{lrrrr}
\toprule
Stratum & $n$ & Official J\&F & Ours \texttt{dw012} & $\Delta$ \\
\midrule
All & 5822 & \second{54.61} & \second{55.91} & +1.30 \\
Small target & 1459 & 33.65 & 35.91 & \best{+2.26} \\
Fast motion & 1457 & 49.57 & 50.94 & +1.37 \\
Occlusion or disappearance & 1459 & 31.29 & 33.22 & +1.92 \\
Distractor-heavy & 1460 & \best{55.95} & \best{57.97} & \second{+2.02} \\
Reasoning query & 2475 & 51.74 & 53.50 & +1.76 \\
Hard union & 4699 & 50.47 & 52.07 & +1.59 \\
\bottomrule
\end{tabular}
\end{table}

\textbf{Global alignment is more stable than token-level alignment.}
We further evaluate whether DINOv2 token-level teacher support can improve fine-grained video segmentation. As shown in Table~\ref{tab:global_token_ablation}, token-only alignment achieves 56.26 J\&F and global+token alignment achieves 56.07 J\&F, both lower than the global-only REPA-Global objective with 57.41 J\&F. Although DINOv2 provides strong patch-level semantic features, directly imposing dense token-level constraints may over-regularize the student visual tokens and interfere with language-conditioned target selection. Therefore, we adopt global representation alignment as a more stable training-time prior.

\textbf{The current evidence supports a mechanism-oriented interpretation.}
Although we do not include a separate feature-trajectory probe, the current evidence is consistent with the intended representation-level mechanism. The KD on/off comparison shows that the gain comes from the teacher-prior term rather than only from the training recipe. The global-versus-token alignment ablation further shows that a compact global constraint is more effective than dense token-level imitation for language-conditioned video segmentation. Together with the temporal stability proxy in the appendix, these results suggest that REPA-Global acts as a representation-level temporal regularizer, rather than only as a mask-head adjustment.

\begin{figure}[t]
  \centering
  \includegraphics[width=1\linewidth]{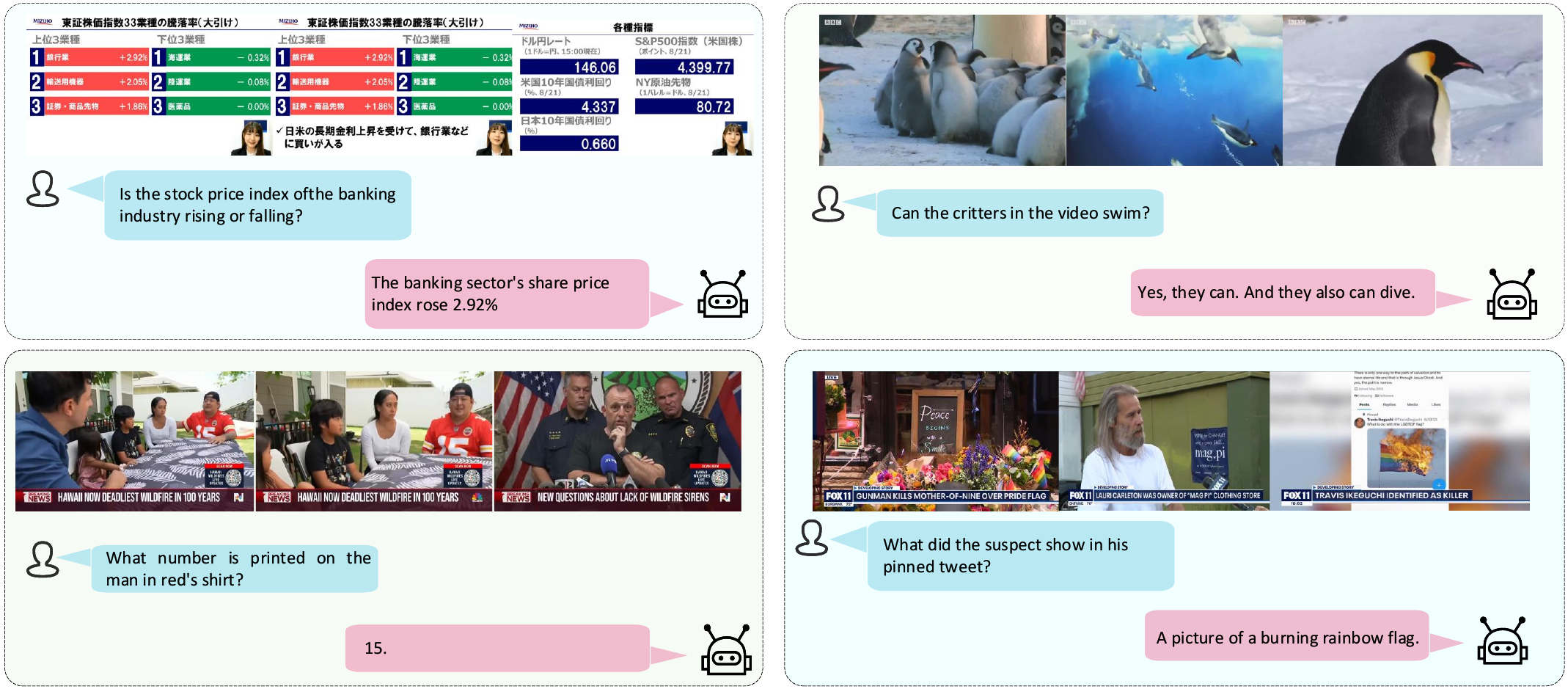}
  \caption{Qualitative visual video QA examples. Each panel shows the input video, the question, and the model response, illustrating that the injected priors do not noticeably degrade multimodal reasoning on representative cases.}
  \label{fig:qa_examples}
\end{figure}

\begin{table}[!t]
\centering
\tabfs
\tabstretch
\setlength{\tabcolsep}{4.8pt}
\caption{
Unified evaluation across image grounding, video segmentation, and MMBench under the InternVL3-2B + SAM2 backbone. We report RefCOCO accuracy, ReVOS and MeVIS U J\&F, and MMBench score. Higher is better.
}
\label{tab:kdsweep_full}
\begin{tabular}{lcccc}
\toprule
Run ID & RefCOCO & ReVOS & MeVIS U & MMBench \\
\midrule
baseline & 81.42\% & 56.06 & 53.92 & \second{0.7876} \\
\texttt{dw012} & 81.61\% & \best{57.41} & \second{55.12} & \best{0.7878} \\
\texttt{dw008} & \best{81.73\%} & \second{56.88} & 54.51 & \best{0.7878} \\
\texttt{dw020} & \second{81.68\%} & 56.87 & \best{55.24} & \best{0.7878} \\
\bottomrule
\end{tabular}
\end{table}
\subsection{Default Configuration and General Grounding}
\label{sec:general_grounding}

Table~\ref{tab:kdsweep_full} summarizes the default configuration across image grounding, video segmentation, and MMBench. The default \texttt{dw012} setting gives the strongest ReVOS performance and competitive MeVIS U transfer, while RefCOCO and MMBench remain close to the baseline. Additional MME, POPE, TextVQA, RefCOCO-series, and distillation-weight sensitivity results are reported in the appendix. These results indicate that the physics-inspired prior improves video consistency without causing a clear collapse in image-level grounding or representative general VLM capability.

\subsection{Limitations and Failure Cases}
\label{sec:limitations}

PhysMLLMs still has several limitations. As shown in the appendix failure cases, dense similar distractors, persistent occlusion, and extremely small targets can still cause incorrect localization or identity ambiguity. Although Table~\ref{tab:revos_stratified} shows gains on small-target and occlusion/disappearance strata, their absolute J\&F scores remain much lower than the full validation average. This indicates that these scenarios are not fully solved.

The limitation mainly comes from the global nature of REPA-Global. The injected physics-inspired prior encourages object permanence, spatial continuity, and temporal coherence through a soft representation-level regularizer, but it does not explicitly model physical forces, 3D dynamics, object interactions, or query-specific target identity. Direct feature-trajectory verification also remains future work. These limitations suggest that query-aware or mask-aware prior injection may further improve fine-grained video grounding.

\section{Conclusion}
\label{sec:conclusion}

PhysMLLMs is a training-stage prior injection strategy for Video-MLLM-based language-guided video segmentation. It targets common temporal failures, including jitter, drift, and identity switches that often appear under occlusion and distractors. The key idea is to inject a physics-inspired spatial continuity prior by aligning the student global visual representation with a frozen DINOv2 teacher through REPA-Global. Teacher supervision is delivered through an offline embedding cache, which makes distillation scalable and keeps inference unchanged because the teacher is removed at test time. We further combine scheduled distillation weighting with calibrated parameter-efficient fine-tuning, which improves training stability and helps preserve general grounding ability while improving mask quality and cross-frame consistency.

The complexity-stratified ReVOS analysis further shows that the gains are concentrated in difficult video scenarios, including small targets, fast motion, occlusion or disappearance, distractor-heavy scenes, and reasoning queries. This supports our motivation that the injected prior mainly addresses temporal identity and spatial continuity. Meanwhile, the RefCOCO-series and general VLM results indicate that the proposed training-time prior preserves image-level grounding and representative multimodal capability.

At the same time, the current evidence mainly supports the proposed mechanism through controlled ablations, challenging-scenario analysis, and temporal stability proxies. Direct feature-trajectory analysis remains future work, especially for query-specific cases with persistent occlusion, extremely small targets, or dense similar distractors.

\Acknowledgements{This work was supported by the National Natural Science Foundation of China (Grants No. 62227807 and U24B20186). This work was also supported by the Supercomputing Center of Lanzhou University.}


\section*{Appendix A. Additional Experimental Results}
\label{app:additional_results}

This appendix provides supplementary quantitative results that support the main experimental conclusions, including general VLM capability checks, RefCOCO-series referring image segmentation comparisons, and distillation-weight sensitivity analysis. These results are moved from the main text to keep the core experimental section concise while preserving the detailed evidence for capability preservation and hyperparameter sensitivity.

\subsection*{A.1 General VLM Capability Check}
\label{app:general_vlm}

All rows in Table~\ref{tab:app_general_vlm_check} are based on the same Sa2VA-InternVL3-2B backbone. The listed entries denote checkpoints obtained at different calibration stages, including the official checkpoint, the RefCOCO fine-tuned initialization, ReVOS-calibrated checkpoints, and the ReVOS-calibrated checkpoint with REPA-Global KD.
Table~\ref{tab:app_general_vlm_check} reports additional general VLM capability checks on MME, POPE, and TextVQA. The ReVOS calib it500 + KD dw012 checkpoint achieves an MME total score of 2152.880, which is comparable to the official Sa2VA-InternVL3-2B score of 2151.195. The nearby ReVOS-calibrated checkpoints also remain stable on POPE and TextVQA. These results suggest that the video-oriented calibration path and REPA-Global injection do not cause an obvious collapse in representative general VLM capability.

\begin{table}[!t]
\centering
\tabfs
\tabstretch
\setlength{\tabcolsep}{2.8pt}
\caption{
General VLM capability check on MME, POPE, and TextVQA. All rows use the same Sa2VA-InternVL3-2B backbone and compare checkpoints from different calibration stages rather than different model backbones. Higher is better.
}
\label{tab:app_general_vlm_check}
\begin{tabular}{lccccc}
\toprule
Checkpoint / training stage & MME perception & MME reasoning & MME total & POPE F1 & TextVQA val acc \\
\midrule
Sa2VA-InternVL3-2B official & \best{1630.838} & 520.357 & 2151.195 & 87.421 & \second{76.808} \\
RefCOCO-ft init & \second{1626.126} & \best{536.071} & \best{2162.198} & 87.405 & 76.580 \\
ReVOS calib it200 & 1610.827 & 522.500 & 2133.327 & \best{87.575} & \best{76.876} \\
ReVOS calib it500 & 1602.206 & 527.857 & 2130.064 & \second{87.542} & 76.796 \\
ReVOS calib it500 + KD \texttt{dw012} & 1622.166 & \second{530.714} & \second{2152.880} & -- & -- \\
\bottomrule
\end{tabular}
\end{table}

\subsection*{A.2 RefCOCO-Series Referring Image Segmentation}
\label{app:refcoco}

Table~\ref{tab:app_refcoco_comparison} reports the RefCOCO-series comparison with representative fine-tuned models. PhysMLLMs remains competitive on RefCOCO, RefCOCO+, and RefCOCOg. Since the proposed spatial prior is designed mainly for video consistency, these results are used as non-degradation checks for image-level referring capability.

\begin{table}[!t]
\centering
\tabfs
\tabstretch
\setlength{\tabcolsep}{5pt}
\caption{
Referring image segmentation comparison with representative fine-tuned models. We report validation accuracy on RefCOCO, RefCOCO+, and RefCOCOg. Higher is better.
}
\label{tab:app_refcoco_comparison}
\begin{tabular}{lccc}
\toprule
Model & RefCOCO & RefCOCO+ & RefCOCOg \\
\midrule
LAVT~\cite{yang2022lavtlanguageawarevisiontransformer} & 72.7 & 62.1 & 61.2 \\
GLaMM-7B~\cite{rasheed2024glammpixelgroundinglarge} & 79.5 & 72.6 & 74.2 \\
OMG-LLaVA-7B~\cite{zhang2024omgllavabridgingimagelevelobjectlevel} & 78.0 & 69.1 & 72.9 \\
F-LMM-7B~\cite{wu2025flmmgroundingfrozenlarge} & 76.1 & 65.2 & 68.5 \\
Sa2VA-InternVL3-2B~\cite{yuan2025sa2va} & \second{81.4} & \second{75.7} & \second{80.3} \\
\midrule
PhysMLLMs & \best{81.9} & \best{76.3} & \best{80.4} \\
\bottomrule
\end{tabular}
\end{table}

\subsection*{A.3 Distillation-Weight Sensitivity}
\label{app:kd_sensitivity}

Table~\ref{tab:app_dw_schedule_compact} reports the sensitivity to the peak distillation weight under the same warmup--ramp--hold schedule. Moderate distillation gives the best ReVOS J\&F, while stronger distillation tends to improve the temporal stability proxy with a small cost in mask quality. This supports using dw012 as the default setting in the main experiments.
\begin{table}[!t]
\centering
\tabfs
\tabstretch
\setlength{\tabcolsep}{4.2pt}
\caption{
Sensitivity to the peak distillation weight under a fixed warmup-ramp-hold schedule. The column ``Peak weight $\lambda_{\max}$'' denotes the maximum scheduled KD weight. Higher tIoU mean and lower tIoU variance indicate stronger temporal stability.
}
\label{tab:app_dw_schedule_compact}
\begin{tabular}{lcccc}
\toprule
Setting & Peak weight $\lambda_{\max}$ & ReVOS J\&F $\uparrow$ & tIoU mean $\uparrow$ & tIoU var $\downarrow$ \\
\midrule
\texttt{dw0002} & 0.002 & 56.64 & 0.6461 & 0.0841 \\
\texttt{dw004}  & 0.004 & 56.53 & 0.6452 & 0.0843 \\
\texttt{dw006}  & 0.006 & 56.59 & 0.6396 & 0.0854 \\
\texttt{dw008}  & 0.008 & \second{56.88} & 0.6476 & 0.0839 \\
\texttt{dw010}  & 0.010 & 56.66 & \best{0.6514} & \second{0.0824} \\
\texttt{dw012}  & 0.012 & \best{57.41} & 0.6419 & 0.0854 \\
\texttt{dw015}  & 0.015 & 56.82 & 0.6451 & 0.0868 \\
\texttt{dw020}  & 0.020 & 56.87 & \second{0.6505} & \best{0.0817} \\
\bottomrule
\end{tabular}
\end{table}

\subsection*{A.4 Qualitative Success Cases}
\label{app:qual_success}

\begin{figure*}[!t]
\centering
\includegraphics[width=\textwidth]{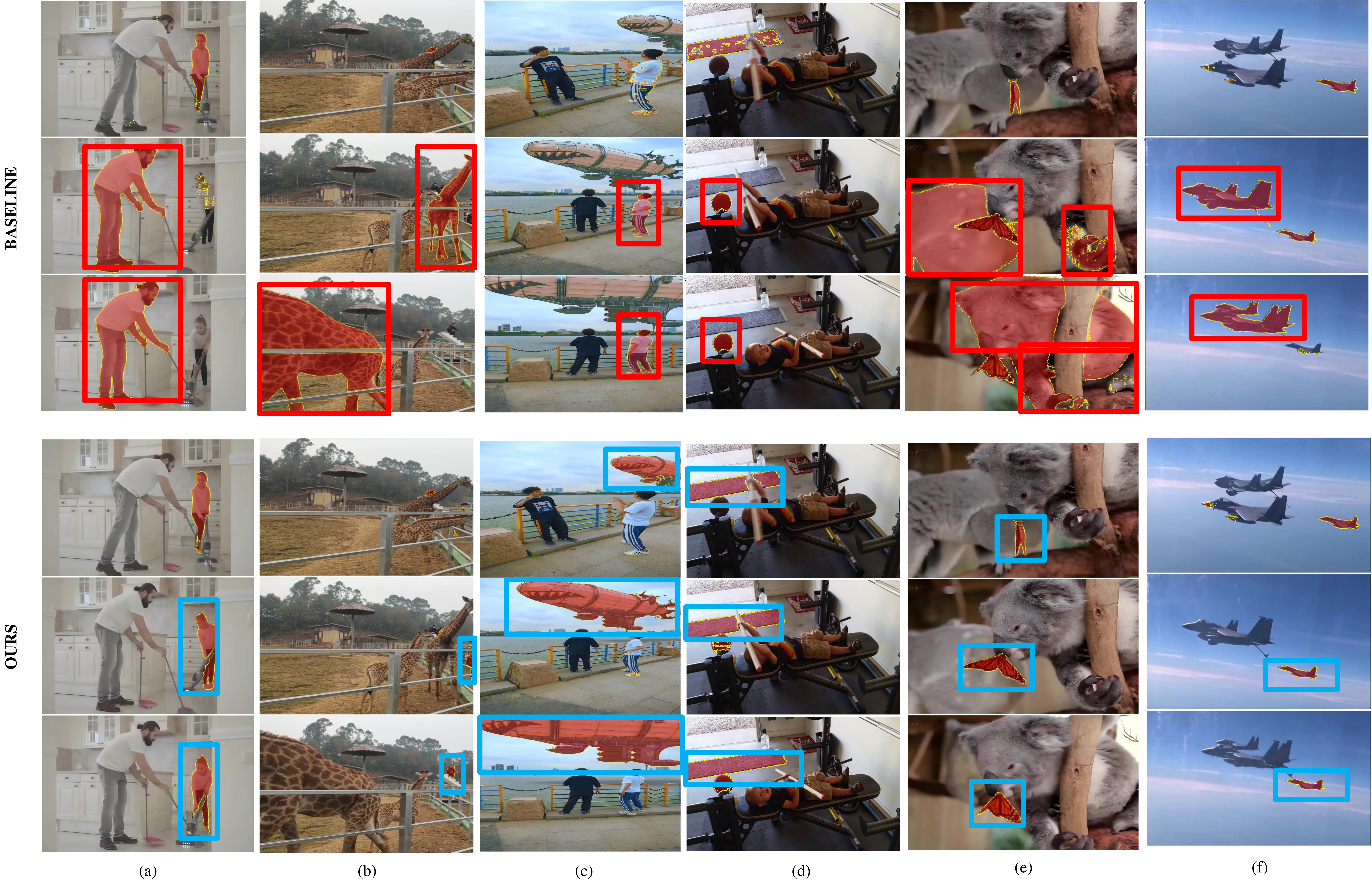}
\caption{
Qualitative comparison on LV-VIS for language-guided video segmentation. The top three rows show the baseline results and the bottom three rows show PhysMLLMs results. Red boxes highlight baseline failures, and blue boxes highlight improved predictions by PhysMLLMs. The segmentation prompts are: 
(a) ``Please segment a person holding a mop.'' 
(b) ``Which creatures are primates? Please answer using segmentation masks.'' 
(c) ``Please segment the object(s) carrying ammunition.'' 
(d) ``Which gray mat(s) have patterns? Please respond with a segmentation mask.'' 
(e) ``Please segment the object fanning its wings on the koala's snout.'' 
(f) ``Please segment the aircraft that is most likely to be in a state of lacking fuel.'' 
Overall, PhysMLLMs improves mask quality and cross-frame consistency under occlusion, distractors, and reasoning-style referring expressions.
}

\label{fig:app_qualitative_success}
\end{figure*}

Figure~\ref{fig:app_qualitative_success} provides additional qualitative comparisons on LV-VIS. These examples cover human-object interaction, multiple similar instances, small or thin objects, and reasoning-style language queries. The baseline often produces unstable masks, such as over-segmented regions under distractors, missing small objects, or identity drift after partial occlusion. PhysMLLMs produces more coherent masks across frames and more accurate target localization in these examples. This supports the quantitative finding that the physics-inspired prior mainly improves temporal consistency and target identity preservation.

\subsection*{A.5 Failure Cases}
\label{app:failure_cases}

\begin{figure*}[!t]
\centering
\includegraphics[width=\textwidth]{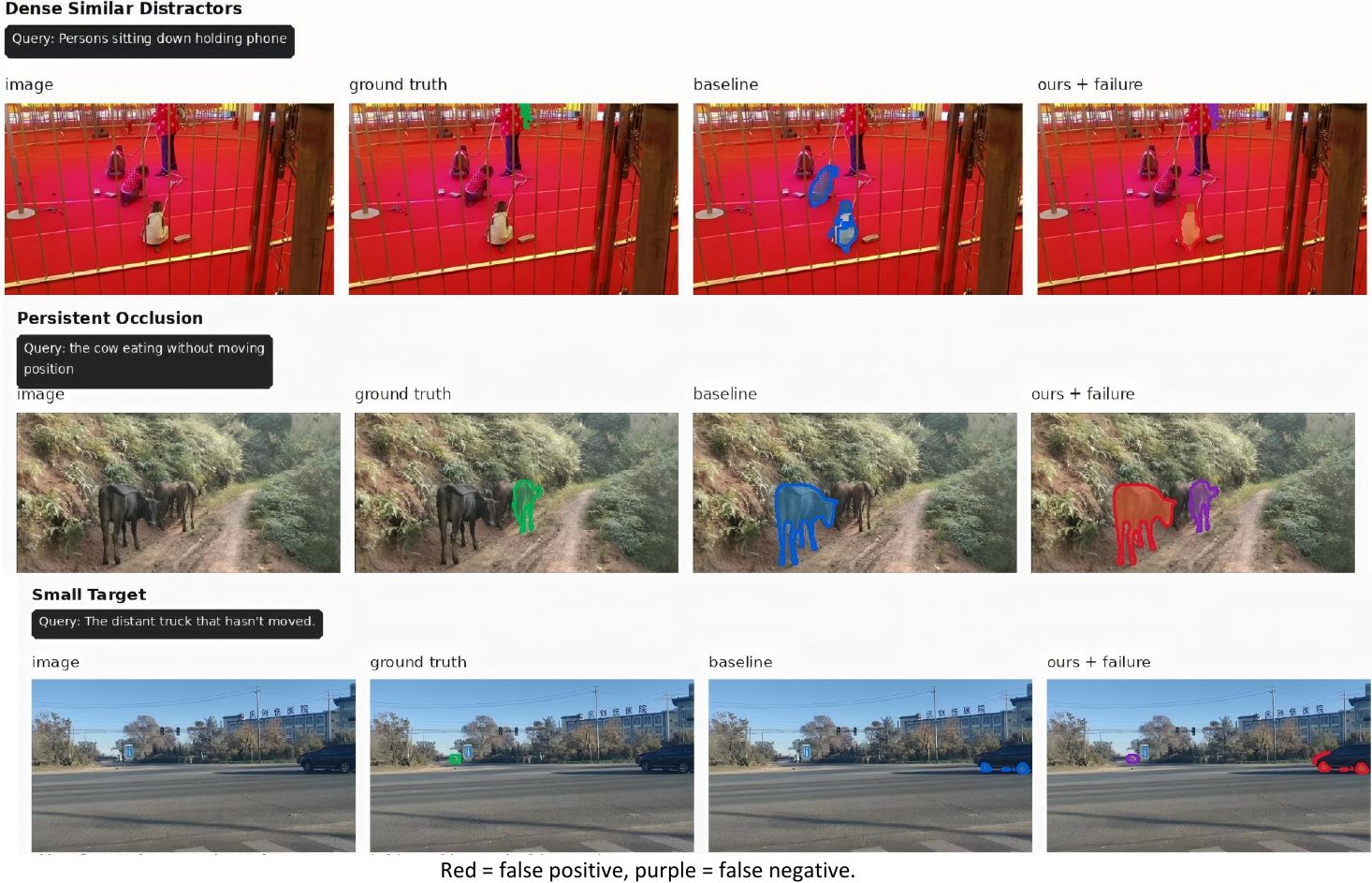}
\caption{
Representative failure cases of PhysMLLMs. Each row shows the input image, ground truth, baseline prediction, and our prediction. The three rows correspond to dense similar distractors, persistent occlusion, and small target cases. Red regions indicate false positives, and purple regions indicate false negatives. These examples show that global prior alignment improves temporal stability but may still be insufficient for query-specific identity disambiguation under highly ambiguous visual conditions.
}
\label{fig:app_failure_cases}
\end{figure*}

Figure~\ref{fig:app_failure_cases} shows representative failure cases of PhysMLLMs. Although the proposed physics-inspired prior improves temporal consistency in many challenging scenarios, it can still fail under dense similar distractors, persistent occlusion, and extremely small targets.

\clearpage


\bibliographystyle{scis}
\bibliography{scis_paper}






\newcommand{\authorbio}[6][l]{%
  \par\medskip\noindent
  \begin{wrapfigure}[#2]{#1}{#3}
    \vspace{-14pt}
    \centering
    \includegraphics[width=#4]{#5}
    \vspace{-6pt}
  \end{wrapfigure}%
  \noindent\hspace{0.2em}#6\par
  \medskip
  \noindent\ignorespaces
}

\end{document}